\documentclass[a4paper,conference]{IEEEtran}
\ifCLASSINFOpdf
  \usepackage[pdftex]{graphicx}
\else
\fi
\usepackage{hyperref}
\hypersetup{
  pdfauthor={Castro et al.},
  pdftitle={PFM for Multiview Gait Recognition},
  pdfsubject={ICPR submission},
  urlcolor=blue,
}

\usepackage[cmex10]{amsmath}
\usepackage{url}

\usepackage[usenames,dvipsnames]{color}

\begin{document}
%
\title{Pyramidal Fisher Motion\\ for Multiview Gait Recognition}

\author{\IEEEauthorblockN{Francisco M. Castro}
\IEEEauthorblockA{IMIBIC --
Dep. of Computing\\ and Numerical Analysis\\
University of Cordoba\\
Cordoba, Spain\\
Email: i92capaf@uco.es}
\and
\IEEEauthorblockN{Manuel J. Mar\'in-Jim\'enez}
\IEEEauthorblockA{IMIBIC --
Dep. of Computing\\ and Numerical Analysis\\
University of Cordoba\\
Cordoba, Spain\\
Email: mjmarin@uco.es}
\and
\IEEEauthorblockN{Rafael Medina-Carnicer}
\IEEEauthorblockA{IMIBIC --
Dep. of Computing\\ and Numerical Analysis\\
University of Cordoba\\
Cordoba, Spain\\
Email: rmedina@uco.es}
}


%


\maketitle

\begin{abstract}
The goal of this paper is to identify individuals by analyzing their gait.
Instead of using binary silhouettes as input data (as done in many previous works) we 
propose and evaluate the use of motion descriptors based on densely sampled short-term trajectories.
We take advantage of state-of-the-art people detectors to define custom spatial configurations of the descriptors around the target person. Thus, obtaining a pyramidal representation of the gait motion. 
The local motion features (described by the Divergence-Curl-Shear descriptor~\cite{jain2013cvpr}) extracted on the different spatial areas of the person are combined into a single high-level gait descriptor by using the Fisher Vector encoding~\cite{perronnin2010eccv}. The proposed approach, coined \textit{Pyramidal Fisher Motion}, is experimentally validated on the recent `AVA Multiview Gait' dataset~\cite{avamvg}. 
The results show that this new approach achieves promising results in the problem of gait recognition.
\end{abstract}


%
\IEEEpeerreviewmaketitle

\section{Introduction}

The term \textit{gait} refers to the way each person walks. Actually, humans are good recognizing people at a distance thanks to their gait~\cite{cutting1977gait}, what provides a good (non invasive) way to identify people without requiring their cooperation, in contrast to other biometric approaches as iris or fingerprint analysis. One potential application of gait recognition is video surveillance, where it is crucial to identify dangerous people without their cooperation.
Although great effort has been put into this problem in recent years~\cite{hu2004survey}, it is still far from solved.

Popular approaches for gait recognition require the computation of the binary silhouettes of people~\cite{martin2012eccv}, usually, by applying some background segmentation technique. However, this is a clear limitation in presence of dynamic backgrounds and/or non static cameras, where noisy segmentations are obtained.
To deal with these limitations, we propose the use of descriptors based on the local motion of points. These kind of descriptors have become recently popular in the field of human action recognition~\cite{wang2011cvpr}. The main idea is to build local motion descriptors from densely sampled points. Then, these local descriptors are aggregated into higher level descriptors by using histogram-based techniques (e.g. Bag of Words~\cite{Sivic03}).

\begin{figure}[th]
\includegraphics[width=0.48 \textwidth]{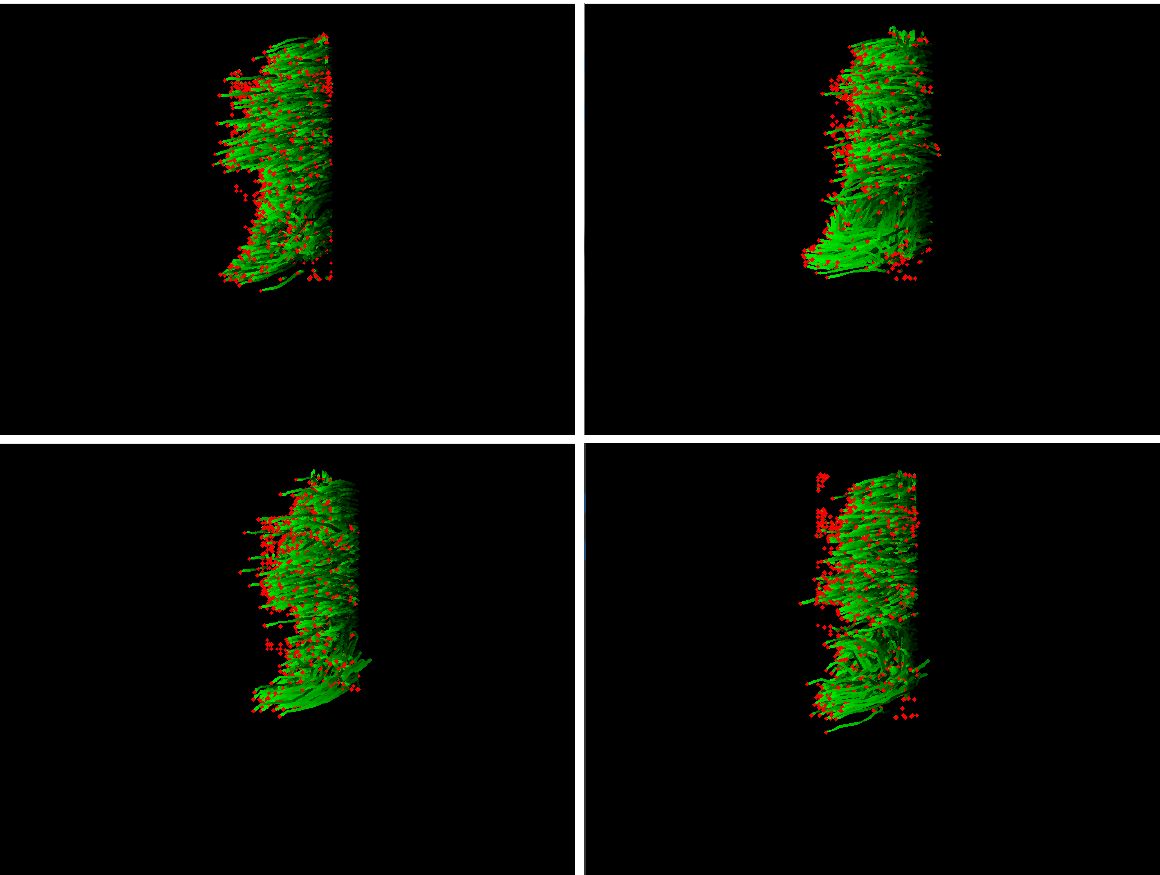}
\caption{\textbf{Who are they?} The goal of this work is to identify people by using their gait. 
We build the Pyramidal Fisher Motion descriptor from trajectories of points. We represent here the gait motion of four different subjects.}
\label{fig:teaser}
\end{figure}

Therefore, our research question is: \textit{could we identify people by using only local motion features as represented in Fig.~\ref{fig:teaser}?} We represent in Fig.~\ref{fig:teaser} the local trajectories of image points belonging to four different people. Our goal is to use each set of local trajectories to build a high-level descriptor that allows to identify individuals. In this paper we introduce a new gait descriptor, coined \textit{Pyramidal Fisher Vector}, that combines the potential of recent human action recognition descriptors with the rich representation provided by Fisher Vectors encoding~\cite{perronnin2010eccv}. A thorough experimental evaluation is carried out on the recent `AVA Multiview Gait'  dataset showing that our proposal contributes to the challenging problem of gait recognition by using a modern computational approach.

This paper is organized as follows. After presenting the related work, we describe our proposed framework for gait recognition in Sec.~\ref{sec:methods}. Sec.~\ref{sec:expers} is devoted to the experimental results. And, finally, the conclusions and future work are presented in Sec.~\ref{sec:conclusions}.

\begin{figure*}[th]
\centering
 \includegraphics[width=0.98 \textwidth]{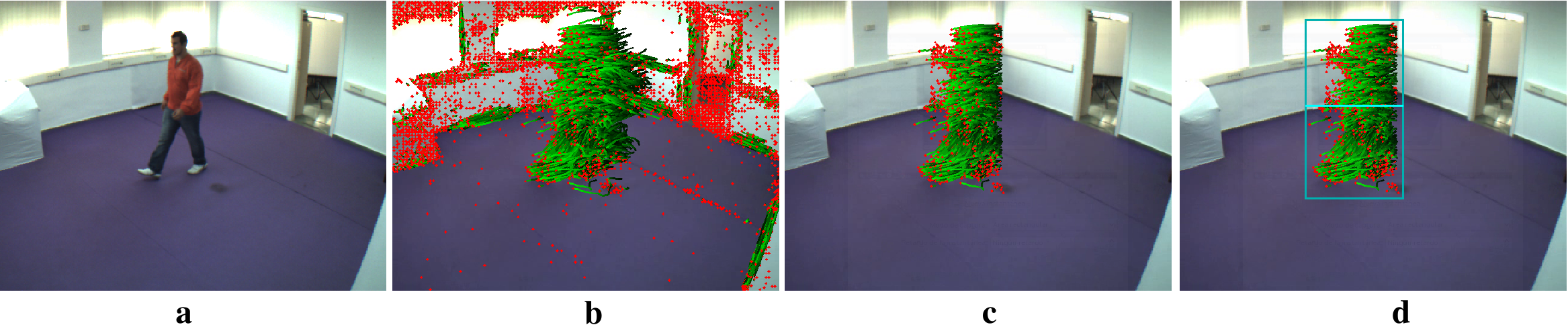}
 \caption{\textbf{Pipeline for gait recognition.} a) The input is a sequence of video frames. b) Densely sampled points are tracked. c) People detection helps to remove trajectories not related to gait. d) A spatial grid is defined on the person bounding-box, so features are spatially grouped to compute a descriptor per cell. Then, those descriptors are concatenated into a single descriptor.}
\label{fig:pipeline}
\end{figure*}

\subsection{Related work}\label{subsec:relwork}
Many research papers have been published in recent years tackling the problem of human gait recognition. For example, in \cite{hu2004survey} we can find a survey on this problem summarizing some of the most popular approaches. Some of them use explicit geometrical models of human bodies, whereas others use only image features. A sequence of binary silhouettes of the body is adopted in many works as input data.
%
%
In this sense, the most popular silhouette-based gait descriptor is the called Gait Enery Image (GEI)~\cite{han2006gei}. The key idea is to compute a temporal averaging of the binary silhouette of the target subject.
Liu et al.~\cite{liu2012icpr}, to improve the gait recognition performance, propose the computation of HOG descriptors from popular gait descriptors as the GEI and the Chrono-Gait Image (CGI).
In \cite{martin2012icpr}, the authors try to find the minimum number of gait cycles needed to carry out a successful recognition by using the GEI descriptor. 
Martin-Felez and Xiang~\cite{martin2012eccv}, using GEI as the basic gait descriptor, propose a new ranking model for gait recognition. This new formulation of the problem allows to leverage training data from different datasets, thus, improving the recognition performance.
In \cite{akae2012cvpr}, Akae et al. propose a temporal super resolution approach to deal with low frame-rate videos for gait recognition. They achieve impressive results by using binary silhouettes of people at a rate of 1-fps.

On the other hand, human action recognition (HAR) is related to gait recognition in the sense that the former also focuses on human motion, but tries to categorize such motion into categories of actions as \textit{walking, jumping, boxing}, etc. In HAR, the work of Wang et al.~\cite{wang2011cvpr} is a key reference. They introduce the use of short-term trajectories of densely sampled points for describing human actions, obtaining state-of-the-art results in the HAR problem. The dense trajectories are described with the Motion Boundary Histogram. Then, they describe the video sequence by using the Bag of Words (BOW) model~\cite{Sivic03}. Finally, they use a non-linear SVM with $\chi^2$-kernel for classification.
In parallel, Perronnin and Dance~\cite{perronnin2007cvpr} introduced a new way of histogram-based encoding for sets of local descriptors for image categorization: the Fisher Vector (FV) encoding. In FV, instead of just counting the number of occurrences of a visual word (i.e. quantized local descriptor) as in BOW, the concatenation of gradient vectors of a Gaussian Mixture is used. Thus, obtaining a larger but richer representation of the image.

Borrowing ideas from the HAR and the image categorization communities, we propose in this paper a new approach for gait recognition that combines low-level motion descriptors, extracted from short-term point trajectories, with a multi-level gait encoding based on Fisher Vectors: the \textit{Pyramidal Fisher Motion} (PFM) gait descriptor.
We have discovered at submission time of this paper a very recent publication that shares some of our ideas, the work of Gong et al.~\cite{gong2013fisher}.
It is similar to ours in the sense that they propose a method that uses dense local spatio-temporal features and a Fisher-based representation rearranged as tensors. However, there are some significant differences: \textit{(i)} instead of using all the local features available in the sequence, we use a person detector to focus only on the ones related to the target subject; \textit{(ii)} the information provided by the person detector enables a richer representation by including coarse geometrical information through a spatial grid defined on the person bounding-box; and, \textit{(iii)} instead of dealing with a single camera viewpoint, we integrate in our system several camera viewpoints. 

\section{Proposed framework}\label{sec:methods}
%

In this section we present our proposed framework to address the problem of gait recognition. 
Fig.~\ref{fig:pipeline} summarizes the pipeline of our approach.
We start by computing local motion descriptors from tracklets of densely sampled points on the whole scene (Fig.~\ref{fig:pipeline}.b -- Sec.~\ref{subsec:features}). Since, we do not assume a static background, we run a person detector to remove the point trajectories that are not related to people (Fig.~\ref{fig:pipeline}.c --Sec.~\ref{subsec:detection}). In addition, we spatially divide the person regions to aggregate the local motion descriptors into mid-level descriptors (Fig.~\ref{fig:pipeline}.d --Sec.~\ref{subsec:fv}). Finally, a discriminative classifier is used to identify the subjects (Sec.~\ref{subsec:svm}).
 

\subsection{Motion-based features}\label{subsec:features}
The first step of our pipeline is to compute densely sampled trajectories. Those trajectories are computed by following the approach of Wang et al.~\cite{wang2011cvpr}. 
Firstly, dense optical flow $F = (u_t,v_t)$ is computed~\cite{Farneback03} on a dense grid (i.e. step size of 5 pixels and over 8 scales). Then, each point $p_t = (x_t, y_t)$ at frame $t$ is tracked to the next frame by
median filtering as follows:
\begin{equation}
   p_{t+1} = (x_{t+1}, y_{t+1}) = (x_t, y_t) + (M * F)|_{(\bar{x_t},\bar{y_t})} 
\end{equation}
where $M$ is the kernel of median filtering and $(\bar{x_t},\bar{y_t})$ is
the rounded position of $p_t$. To minimize drifting effect, the tracking is limited to $L$ frames. We use $L=15$ as in~\cite{jain2013cvpr}. As a postprocessing step, noisy and uninformative trajectories (e.g. excessively short or showing sudden large displacements) are removed.

Once the local trajectories are computed, they are described with the Divergence-Curl-Shear (DCS) descriptor proposed by Jain et al.~\cite{jain2013cvpr}, which is computed as follows:

\begin{equation}\label{eq:DCS}
\left\{
\begin{aligned}
	\text{div}(p_t)  &=& \frac{\partial u(p_t)}{\partial x} + \frac{\partial v(p_t)}{\partial y} \quad \\
	\text{curl}(p_t) &=& \frac{-\partial u(p_t)}{\partial y} + \frac{\partial v(p_t)}{\partial x} \quad \\
	\text{hyp}_1(p_t) &=& \frac{\partial u(p_t)}{\partial x} - \frac{\partial v(p_t)}{\partial y} \quad \\
	\text{hyp}_2(p_t) &=& \frac{\partial u(p_t)}{\partial y} + \frac{\partial v(p_t)}{\partial x} \quad \\	
\end{aligned}
\right.
\end{equation}

As described in \cite{jain2013cvpr}, the divergence is related to axial motion, expansion and scaling effects, whereas the curl is related to rotation in the image plane. From the hyperbolic terms ($\text{hyp}_1,\text{hyp}_2$), we can compute the magnitude of the shear as:

\begin{equation}\label{eq:shear}
  \text{shear}(p_t) = \sqrt{\text{hyp}_1^2(p_t) + \text{hyp}_2^2(p_t)}
\end{equation}

\subsection{People detection and tracking}\label{subsec:detection}
We follow a tracking-by-detection strategy as in~\cite{Eichner2012ijcv}: we detect full bodies with the detection framework of Felzenszwalb et al.~\cite{felzenszwalb10pami}; and, then, we apply the clique partitioning algorithm of Ferrari et al.~\cite{Ferrari01} to group detections into tracks.
Short tracks with low-scored detections are considered as false positives and are discarded for further processing. In addition, to remove false positives generated by static objects, we measure the displacement of the detection along the sequence. Thus, discarding those tracks showing a static behaviour.

The tracks finally kept are used to filter out the trajectories that are not related to people: we only keep the trajectories that pass through, at least, one bounding-box of any track. In this way, we can focus on the trajectories that should contain information about the gait.

\subsection{Pyramidal Fisher Motion}\label{subsec:fv}
\noindent \textbf{Fisher Vector encoding.} %
As described above, our low-level features are based on motion properties extracted from person-related local trajectories. In order to build a person-level gait descriptor, we need to summarize the local features. We propose here the use of Fisher Vectors (FV) encoding~\cite{perronnin2010eccv}.

The FV, that can be seen as an extension of the Bag of Words (BOW) representation~\cite{Sivic03}, builds on top of a Gaussian Mixture Model (GMM), where each Gaussian corresponds to a visual word. Whereas in BOW, an image is represented by the number of occurrences of each visual word, in FV an image is described by a gradient vector computed from a generative probabilistic model.

Assuming that our local motion descriptors $\{x_t \in R^D,t=1 \ldots T\}$ of a video $V$ are generated independently by a GMM $p(x|\lambda)$ with parameters $\lambda=\{w_i, \mu_i, \Sigma_i , i=1 \ldots N\}$, we can represent $V$ by the following gradient vector~\cite{perronnin2007cvpr}:
\begin{equation}\label{eq:fv}
G_\lambda (V) = \frac{1}{T} \sum_{t=1}^{T}{\nabla_\lambda \log p(x_t|\lambda)}
\end{equation}

Following the proposal of~\cite{perronnin2010eccv}, to compare two videos $V$ and $W$, a natural kernel on these gradients is the Fisher Kernel: $K(V,W) = G_\lambda(V)^T F_\lambda^{-1} G_\lambda(W)$, where $F_\lambda$ is the Fisher Information Matrix~\cite{jaakkola1999nips}.

As $F_\lambda$ is symmetric and positive definite, it has a Cholesky decomposition $F_\lambda^{-1}=L_\lambda^T L_\lambda$, and 
$K(V,W)$ can be rewritten as a dot-product between normalized vectors $\Gamma_\lambda$ with: $\Gamma_\lambda(V)=L_\lambda G_\lambda(V)$. Then, $\Gamma_\lambda(V)$ is known as the Fisher Vector of video V.
As stated in \cite{perronnin2010eccv}, the capability of description of the FV can be improved by applying it a signed square-root followed by L2 normalization. So, we adopt this finding for our descriptor.

The dimensionality of FV is $2ND$, where $N$ is the number of Gaussians in the GMM, and $D$ is the dimensionality of the local motion descriptors $x_t$. For example, in our case, the dimensionality of the local motion descriptors is $D=318$, if we use $N=100$ Gaussians, then, the FV would have 63600 dimensions.
In this paper, we will use the term \textit{Fisher Motion} (FM) to refer to the FV computed on a video from low-level motion features.

\begin{figure}[t]
\includegraphics[width=0.45 \textwidth]{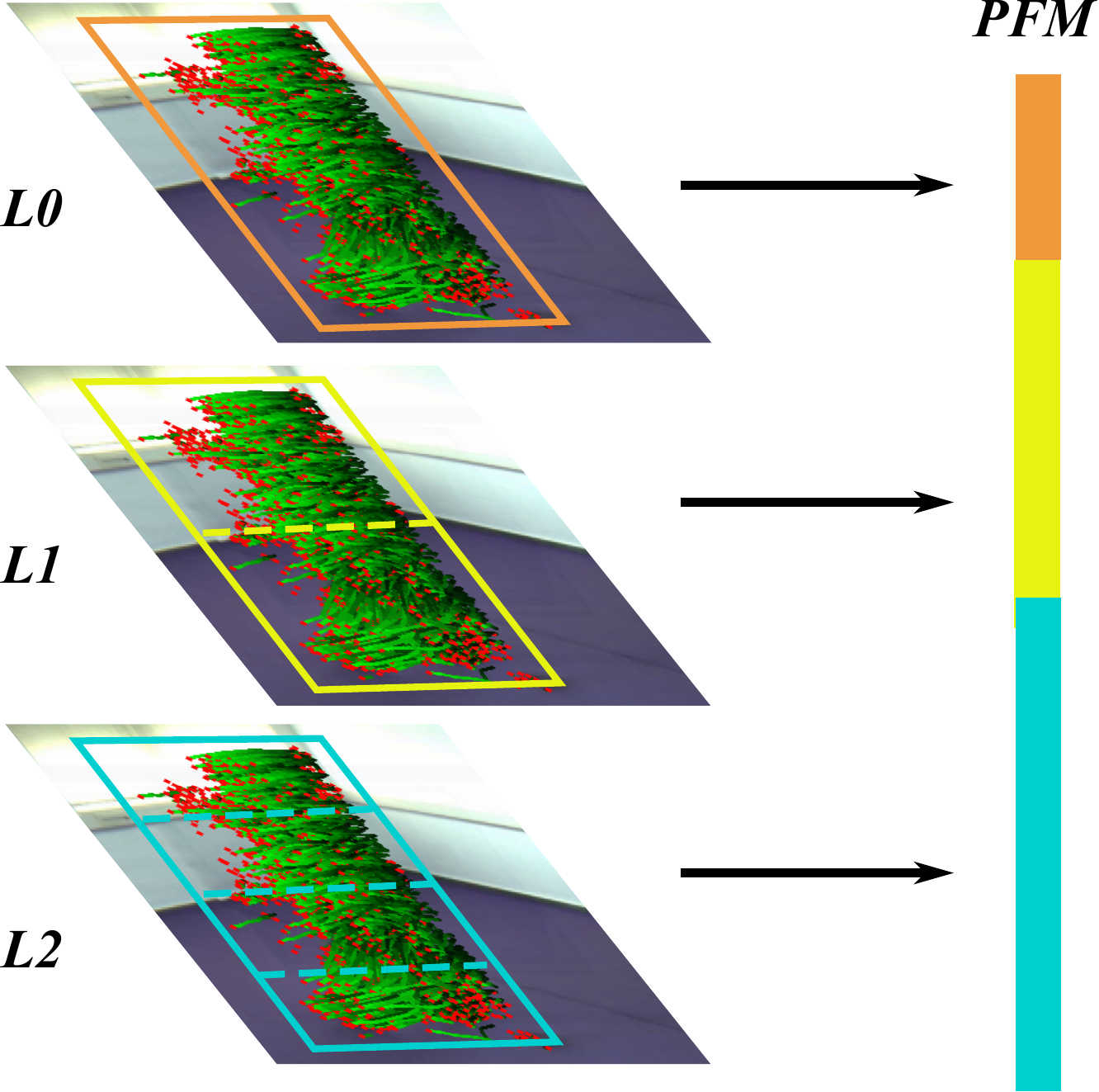}
\caption{\textbf{Pyramidal Fisher Motion descriptor.} Fisher Vector encoding is used at each spatial region on the Divergence-Curl-Shear descriptors computed on the dense trajectories enclosed in the person bounding-box. All Fisher Vectors are concatenated to obtain the final PFM gait descriptor.}
\label{fig:pfm}
\end{figure}

\noindent \textbf{Pyramidal representation.} %
We borrow from \cite{Marin2012paaa} the idea of building a pyramidal representation of the gait motion.
Since each bounding-box covers the whole body of a single person, we propose to spatially divide the BB into cells. 
Then, a Fisher vector is computed inside each cell of the spatio-temporal grid. 
We can build a pyramidal representation by combining different grid configurations. Then, the final feature vector, 
used to represent a time interval, is computed as the concatenation of the cell-level Fisher vectors from all the levels of the pyramid. This idea is represented in Fig.\ref{fig:pfm}, where each colored segment of the PFM descriptor comes from a different level of the pyramid.

\begin{figure*}[th]
\begin{center}
  \includegraphics[width=0.98 \textwidth]{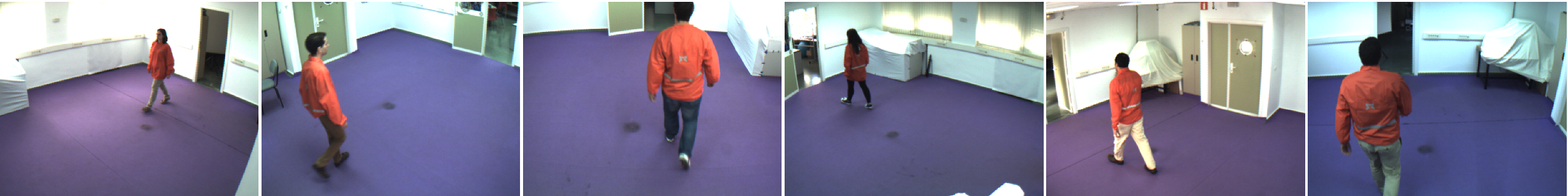}%
 \caption{\textbf{AVAMVG dataset.} Different people recorded from six camera viewpoints. The dataset contains both female and male subjects performing different trajectories through the indoor scenario. Note that cameras 3rd and 6th (from left to right) are prone to show people partially occluded.}
 \label{fig:dataset}
\end{center}
\end{figure*}

\subsection{Classification}\label{subsec:svm}
The last stage of our pipeline is to train a discriminative classifier to distinguish between the different human gaits. 
Since, this is a multiclass problem, we train $P$ binary linear Support Vector Machines (SVM)~\cite{osuna1997svm} (as many as different people) in a \textit{one-vs-all} strategy. 
Although the $\chi^2$ kernel~\cite{vedaldi12pami} is a popular choice for BOW-based descriptors, a linear kernel is typically enough for FV, due to the rich feature representation that it provides. 

\subsection{Implementation details}\label{subsec:code}
For people detection, we use the code published by the authors of \cite{felzenszwalb10pami}.
For computing the local motion features, we use the code published by the authors of \cite{jain2013cvpr}. The Fisher Vector encoding and the classification is carried out by using the code included in the library VLFeat~\footnote{%
VLFeat 0.9.17 is available at \url{http://www.vlfeat.org/}
}.%

\section{Experimental Results}\label{sec:expers}
%
We carry out diverse experiments in order to validate our approach. 
With these experiments we try to answer the following questions: a) \textit{is the combination of trajectory-based features with FV a valid approach for gait recognition?}; b) \textit{can we learn different camera viewpoints in a single classifier?}; c) \textit{can we improve the recognition rate by spatially dividing the human body region?}; d) \textit{what is the effect of using PCA-based dimesionally reduction on the recognition performance?}; and, e) \textit{can the proposed model generallize well on unrestricted walk trajectories?}

\subsection{Dataset}\label{subsec:dataset}
We perform our experiments on the ``AVA Multi-View Dataset for Gait Recognition'' (AVAMVG)~\cite{avamvg}. 
In AVAMVG 20 subjects perform 10 walking trajectories in an indoor environment. 
Each trajectory is recorded by 6 color cameras placed around a room that is crossed by the subjects during the performance.
Fig.~\ref{fig:dataset} shows the scenario from the six available camera viewpoints. Note that depending on the viewpoint and performed trajectory, people appear at diverse scales, even showing partially occluded body parts. In particular, the 3rd and 6th camera viewpoints represented in Fig.~\ref{fig:dataset} are more likely to show partially visible bodies most of the time than the other four cameras. Therefore, in our experiments, and without loss of generality, we will use only four cameras (i.e. $\{1, 2, 4, 5\}$).
Trajectories 1 to 3 follow a linear path, whereas the remaining seven trajectories are curved. The released videos have a resolution of $640 \times 480$ pixels. Each video has around 375 frames, where only approximately one third 
of the frames contains visible people.


\subsection{Experimental setup}\label{subsec:expersetup}
%

Since we have multiple viewpoints of each \textit{instance} (i.e. pair subject--trajectory), we assign a single label to it by majority voting on the viewpoints. This approach helps to deal with labels wrongly assigned to individual viewpoints. Note that instead of training an independent classifier (see Sec.~\ref{subsec:svm}) per camera viewpoint, we train a single classifier with samples obtained from different camera viewpoints, allowing the classifier to learn the relevant gait features of each subject from multiple viewpoints.
In order to increase the amount of training samples, we generate their \textit{mirror} sequences, thus, doubling the samples during learning. 
%
%

We describe below the different experiments performed to give answer to the questions stated at the beginning of this section.

\noindent \textbf{Experiment A: baseline.} %
We use the popular Bag of Words approach (BOW)~\cite{Sivic03} as baseline, which is compared to our approach. For this experiment, we use trajectories 1, 2 and 3 (i.e. straight path). We use a leave-one-out strategy on the trajectories (i.e. two for training and one for test). We sample dictionary sizes in the interval $[500,4000]$ for BOW~\footnote{%
Larger dictionary sizes for BOW did not show any significative improvement. In contrast, the computational time increased enormously. 
}%
, and in the interval $[50,200]$ for PFM. Both BOW and PFMs have a single level with two rows and one column (i.e. concatenation of two descriptors: half upper-body and half lower-body). 

\noindent \textbf{Experiment B: half body features.} %
Focusing on PFM, we compare four configurations of the PFM on trajectories 1, 2 and 3: a) no spatial partition of the body; b) using only the top half of the body; c) using only the bottom half of the body; and, d) using the concatenation of the top and bottom half of the body. 

\noindent \textbf{Experiment C: dimensionality reduction.} %
Since the dimensionallity of PFM is typically large, we evaluate in this experiment the impact of dimensionality reduction on the final recognition performance. We run Principal Component Analysis (PCA) both on the original low-level features (318 dimensions), and on the PFM vectors. We use the PFM descriptor, as in experiment A, on trajectories 1, 2 and 3.

\noindent \textbf{Experiment D: training on straight paths and testing on curved paths.} %
In this experiment, we use the PFM descriptor as in experiment A. We use trajectories 1 to 3 for training, and trajectories 4, 7 and 10 for testing. Note that in the latter sequences, the subjects perform curved trajectories, thus, changing their viewpoint (with regard to a given camera). 

\subsection{Results}
We present here the results of the experiments described above.

\begin{table*}[th] 
\renewcommand{\arraystretch}{1.5}
\caption{\textbf{Comparison of recognition results.} Each entry contains the percentage of correct recognition in the multiview setup and, in parenthesis, the recognition per single view. Each row corresponds to a different configuration of the gait descriptor. 
$K$ is the GMM size used for FM. Best results are marked in bold.
(See main text for further details.)}
\label{tab:resultsA}
\centering
\begin{tabular}{l|c| c c c c}
 \hline
 \textit{Experiment} & $K$ & \textit{Trj=1+2} & \textit{Trj=1+3} & \textit{Trj=2+3} & \textit{Avg}\\
 \hline
  BOW       & 4000 & 95 (78.8) & 85 (62.5) & 100 (84.4) & 93.3 (75.2) \\ 
  PFM-FB    & 150  & 100 (98.8)  &  100 (95)  & 100 (100) & 100 (97.9) \\ 
  PFM-H1     & 150 & 100 (95)  & 100 (87.5)   & 100 (97.5) & 100 (93.3) \\
  PFM-H2     & 150 & 100 (97.5) & 95 (93.8)   & 100 (97.5) & 98.3 (96.3) \\
  PFM        & 150 & 100 (98.8) & 100 (96.2)  & 100 (97.5) & 100 (97.5) \\
  PFM+PCAL50 & 150 & 100 (100) & 100 (97.5)  & 100 (98.8)  & 100 (98.8) \\
  PFM+PCAH256& 100 & 100 (100)  & 100 (97.5)   & 100 (98.8) & 100 (98.8)\\
  PFM+PCAL100+PCAH256& 150 & 100 (100) & 100 (97.5) & 100 (98.8) & 100 (98.8)\\
  \hline
  PFM+PCAL50+PCAH256+pyr& 100 & 100 (100) & 100 (97.5) & 100 (98.8) & \textbf{100 (98.8)}\\
 \hline
\end{tabular}
\end{table*}

The results shown in Tab.~\ref{tab:resultsA} correspond to \textit{experiments A, B and C} (see Sec.~\ref{subsec:expersetup}) and have been obtained by training on two of the three  straight trajectories ($\{1,2,3\}$) and testing on the remaining one (e.g. `\textit{Trj=1+2}' indicates training on trajectories \#1 and \#2, then, testing on trajectory \#3).
Therefore, each model is trained with 160 samples (i.e. 20 subjects $\times$ 4 cameras  $\times$ 2 trajectories) and tested on 80 samples.
Each column `\textit{Trj=X+Y}' contains the percentage of correct recognition per partition at instance level (i.e. combining the four viewpoints) and, in parenthesis, at video level; column `\textit{Avg}' contains the average on the three partitions. Column $K$ refers to the number of centroids used for quantizing the low-level features in each FM descriptor.
Row `BOW' corresponds to the baseline approach (see Sec.~\ref{subsec:expersetup}). 
Row `PFM-FB' corresponds to the PFM on the full body (no spatial partitions). 
Rows `PFM-H1'and `PFM-H2' correspond to PFM on the top half and on the bottom half of the body, respectively.
Row `PFM' corresponds to a single-level PFM obtained by the concatenation of the descriptors extracted from both the top and bottom half of the body.
Row `PFM+PCAL50' corresponds to our proposal but reducing with PCA the dimensionality of the low-level motion descriptors to 50 before building our PFM (i.e. final gait descriptor with $K=150$ is $15000$-dimensional).
Row `PFM+PCAH256' corresponds to our proposal but reducing with PCA the dimensionality of our final PFM descriptor to 256 dimensions before learning the classifiers (i.e. final gait descriptor is $256$-dimensional).
Row `PFM+PCAL100+PCAH256' corresponds to our proposal but reducing both the dimensionality of the low-level descriptors and the final PFM descriptor (i.e. final gait descriptor is $256$-dimensional). 
Row `PFM+PCAL50+PCAH256+pyr' corresponds to a two-level pyramidal configuration where the first level has no spatial partitions and the second level is obtained by dividing the bounding box in two parts along the vertical axis, as done previously. In addition, PCA is applied to both the low-level descriptors and the final PFM vector.

\begin{table}[th] 
\renewcommand{\arraystretch}{1.5}
\caption{\textbf{Comparative of recognition results on curved trajectories.} Training on trajectories $1+2+3$. Each column indicates the tested trajectory and each row corresponds to a different configuration of the gait descriptor. $K$ is the GMM size used for FM. Best results are marked in bold.}
\label{tab:resultsB}
\centering
\begin{tabular}{l|c| c c c}
 \hline
 \textit{Experiment} & $K$ & \textit{Test=04} & \textit{Test=07} & \textit{Test=10}\\
 \hline
  PFM         & 150 & 90 (75) & 95 (91.3) & \textbf{95} (81.0)\\  
  PFM+PCAL100 & 150 & 90 (72.5) & 95 (92.5) & 80 (77.2) \\ 
  PFM+PCAL100+PCAH256 & 150 & 90 (73.8) & 95 (90) & 85 (81.1) \\
  \hline
  PFM+PCAL50+PCAH256+pyr & 100 & \textbf{95 (80)} & 90 (88.8) & 85 (82.3)\\
  PFM+PCAL50+PCAH256+pyr & 150 & 90 (75) & 95 (90) & 90 (\textbf{87.3}) \\
  PFM+PCAL100+PCAH256+pyr & 150 & 85 (71.3) & \textbf{95 (92.5)} & 85 (82.3) \\
 \hline
\end{tabular}
\end{table}

The results shown in Tab.~\ref{tab:resultsB} correspond to \textit{experiment D} (see Sec.~\ref{subsec:expersetup}) and have been obtained by training on trajectories $\{1, 2, 3\}$ (all in the same set), and testing on trajectories $\{4, 7, 10\}$ (see corresponding columns). As done in the previous experiments, different configurations of PFM have been evaluated. Each entry of the table contains the percentage of correct recognition in the multiview setup and, in parenthesis, the recognition per video.

\subsection{Discussion}
%
The results presented in Tab.~\ref{tab:resultsA} indicate that the proposed pipeline is a valid approach for gait recognition, obtaining a $100\%$ of correct recognition on the multiview setup. In addition, the FV-based formulation surpasses the BOW-based one, as stated by other authors in the problem of image categorization~\cite{perronnin2010eccv}.
In addition, the large dimensionality of the PFM can be drastically reduced by applying PCA, without worsening the final performance. Actually, reducing the dimensions of the low-level motion descriptors to 100, and the final PFM to 256, allows to achieve a similar recognition rate but decreasing significantly the computational complexity ($\approx \times370$ smaller with $K=150$).

If we focus on the idea of spatially dividing the human body for computing different gait descriptors, the results show that the most discriminant features are localized on the lower-body (row `PFM-H2'), what confirms our intuition (i.e. gait is mostly defined by the motion of the legs). In addition, although in a slight manner (see values in parenthesis), the upper-body features (row `PFM-H1') contribute to the definition of the gait as well.

Focusing on Tab.~\ref{tab:resultsB}, we can observe that PFM generalizes fairly well, as derived from the results obtained when testing on curved trajectories. From the three tested trajectories, the number \#04 resulted to be the hardest when trying to classify per individual cameras (i.e. values in parenthesis). However, the use of the majority voting strategy on the multiview setup clearly contributed to boost the recognition rate (e.g. from $80$ to $95$).

With regard to the use of more than one level in PFM, we can see in Tab.~\ref{tab:resultsA} that similar results are obtained with the single- and two-level configurations. However, in the two-levels case, the number of needed GMMs is lower (i.e. 100 vs 150) than with single-level, as well as the low-level features can be reduced to half size (i.e. 50 vs 100). In addition, in the experiment on the curved trajectories (Tab.~\ref{tab:resultsB}), we can find in many cases an improvement at video level (e.g. $75$ to $80$ in trajectory \#04). Although we tried an additional third level in the pyramid, the recognition rate did not increase.


In addition to the reported experiments, we also experimented with splitting the curved video sequences along the temporal axis to try to find nearly linear trajectories that could be better classified. However, the results did not show any improvement. 

In summary, we can conclude that the proposed PFM allows to identify subjects by their gait by using as basis local motion (i.e. short-term trajectories) and coarse structural information (i.e. spatial divisions on the person bounding-box). Moreover, PFM does not need either segmenting or aligning the gait cycle of each subject as done in previous works.

\section{Conclusions and Future Work}\label{sec:conclusions}
We have presented a new approach for recognizing human gait in video sequences.
Our method builds a pyramidal representation of the human gait based on the combination of densely sampled local features and Fisher vectors: the \textit{Pyramidal Fisher Motion}. 

The results show that PFM allows to obtain a high recognition rate on a multicamera setup: the AVAMVG dataset. In particular, a perfect identification of the individuals is achieved when we combine information from different cameras and the subjects follow a straight path. In addition, our pipeline shows a good behaviour on unconstrained paths, as shown by the experimental results -- the model is trained on subjects performing straight walking trajectories and tested on curved trajectories.
With regard to the PFM configuration, we have observed that it is beneficial to decorrelate (by using PCA) both the low-level motion features and the final PFM descriptor in order to achieve high recognition results and, in turn, decreasing the computational burden at test time -- the classification with a linear SVM is extremely fast on 256-dimensional vectors.
Since we use a person detector to localize the subjects, the proposed system in not restricted to deal with scenarios with static backgrounds. Moreover, the motion features used in this paper can be easily adapted to non static cameras by removing the global affine motion as proposed recently by Jain et al.~\cite{jain2013cvpr}.

In conclusion, PFM enables a new way of tackling the problem of gait recognition on multiple viewpoint scenarios, removing the need of using people segmentation as mostly done so far.

As future work, we intend to evaluate the proposed method on additional multiview datasets that include both people carrying objects and \textit{impostors} (i.e. people external to the learnt subjects). With regard to the latter issue, as we use the continous output of the SVM to decide the identity, we could eventually discard impostors by thresholding such value.

\section*{Acknowledgments}
This work has been partially funded by the Research Projects TIN2012-32952 and BROCA, both financed by FEDER and the Spanish Ministry of Science and Technology. 
We also thank David L\'opez for his help with the setup of the AVAMVG dataset.



\bibliographystyle{IEEEtran}
\bibliography{shortstrings,local,bibAVA}
%
%
%

\end{document}